%% file: neurips_2023.tex
\pdfoutput=1
\documentclass{article}
\usepackage[preprint]{neurips_2023}
\input{others/shortcuts}

\usepackage[table]{xcolor}
\definecolor{blue}{HTML}{0055cc}
\definecolor{red}{HTML}{cc1100}
\definecolor{orange}{HTML}{cc7700}
\definecolor{green}{HTML}{339955}
\definecolor{Highlight}{HTML}{39a58a}
\usepackage{float}
\usepackage[utf8]{inputenc} % allow utf-8 input
\usepackage[T1]{fontenc}    % use 8-bit T1 fonts
\usepackage[pagebackref,breaklinks,colorlinks,linkcolor=purple,citecolor=Highlight]{hyperref}       % hyperlinks
\usepackage{url}            % simple URL typesetting
\usepackage{booktabs}       % professional-quality tables
\usepackage{amsfonts}       % blackboard math symbols
\usepackage{nicefrac}       % compact symbols for 1/2, etc.
\usepackage{microtype}      % microtypography
\usepackage{xcolor}         % colors
\usepackage{amsmath}        % ams math
\usepackage{amssymb}        % ams symbol
\usepackage{lipsum}         % dummy sentences
\usepackage{bbding}         % checkmark
\usepackage{array}          % table column width
\usepackage{subcaption}     % subtable
\usepackage{pdfpages}
\usepackage{natbib}
\usepackage{adjustbox}
\usepackage{tcolorbox}
\usepackage{enumitem}
\usepackage{ulem}
\usepackage{caption}
\setcitestyle{square,numbers,comma,sort}
\usepackage[capitalize]{cleveref}
\crefname{section}{Sec.}{Secs.}
\Crefname{section}{Section}{Sections}
\Crefname{table}{Table}{Tables}
\crefname{table}{Tab.}{Tabs.}

\newlength\savewidth

\newcolumntype{x}[1]{>{\centering\arraybackslash}p{#1pt}}
\newcolumntype{y}[1]{>{\raggedright\arraybackslash}p{#1pt}}
\newcolumntype{z}[1]{>{\raggedleft\arraybackslash}p{#1pt}}

\newcommand{\app}{\raise.17ex\hbox{$\scriptstyle\sim$}}

\definecolor{deemph}{gray}{0.6}

\definecolor{baselinecolor}{gray}{.9}

\newcommand{\cgaphl}[2]{
	\fontsize{8pt}{1em}\selectfont{\textcolor{Highlight}{($\textbf{#1}$\textbf{#2})}}
}

\newcommand{\cgaphlb}[2]{
	\fontsize{8pt}{1em}\selectfont{\textcolor{black}{($\textbf{#1}$\textbf{#2})}}
}

\definecolor{my_red}{HTML}{FE4444}
\definecolor{Highlight}{HTML}{39a58a}  % green
\definecolor{Gray}{gray}{0.95}

\let\originalleft\left
\let\originalright\right
\renewcommand{\left}{\mathopen{}\mathclose\bgroup\originalleft}
\renewcommand{\right}{\aftergroup\egroup\originalright}

\title{ARPO: End-to-End Policy Optimization \\ for GUI Agents with Experience Replay}

\author{%
  Fanbin Lu\textsuperscript{$1$} \quad 
  Zhisheng Zhong \textsuperscript{$1$} \quad 
  Shu Liu \textsuperscript{$2$} \quad 
Chi-Wing Fu\textsuperscript{$1$} \quad 
    Jiaya Jia\textsuperscript{$2,3$} \quad 
  \vspace{0.1cm} \\
  \textsuperscript{$1$}The Chinese University of Hong Kong \quad
  \textsuperscript{$2$}SmartMore \quad \\
  \textsuperscript{$3$}Hong Kong University of Science and Technology \quad
  \vspace{0.1cm} 
}

\definecolor{lightblue}{RGB}{230,245,255}
\definecolor{lightgray}{gray}{0.95}

\begin{document}

\maketitle

\vspace{-10pt}
\begin{figure*}[h!]
\centering
\begin{subfigure}{0.38\linewidth}
    \centering
    \includegraphics[height=3.7cm]{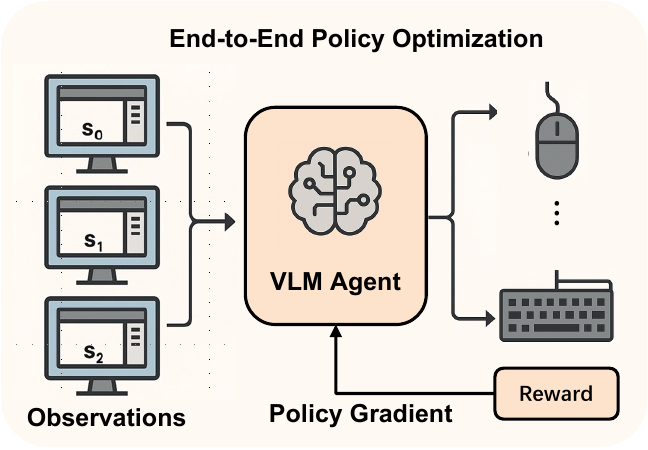}
    \caption{End-to-end GUI agent.}
    \label{fig:teaser_a}
\end{subfigure}
\hfill
\begin{subfigure}{0.25\linewidth}
    \centering
    \includegraphics[height=3.7cm]{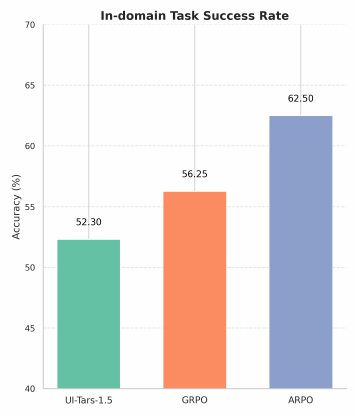}
    \caption{Task completion rate.}
    \label{fig:teaser_b}
\end{subfigure}
\hfill
\begin{subfigure}{0.35\linewidth}
    \centering
    \includegraphics[height=3.7cm]{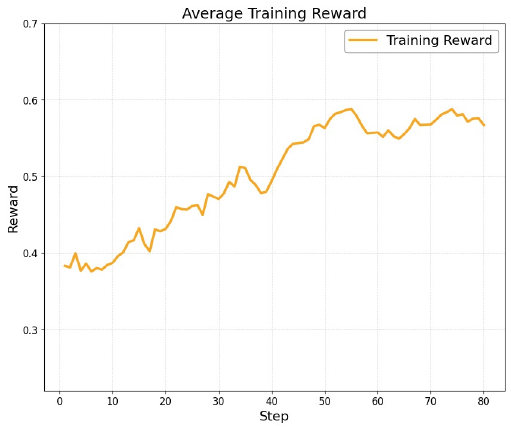}
    \caption{Training rewards.}
    \label{fig:teaser_c}
\end{subfigure}
\caption{\textbf{A}gentic \textbf{R}eplay \textbf{P}olicy \textbf{O}ptimization~(ARPO) enables effective end-to-end policy optimization for GUI agents.
(a) Our vision-language agent processes long-horizon visual observations and interaction histories to generate sequential actions and receive policy gradients from sparse, delayed rewards.
(b) ARPO significantly boosts in-domain task success rates compared to baseline and GRPO-only training.
(c) Average training reward steadily increases, demonstrating improved policy learning and sample efficiency in complex GUI environments.
}
\label{fig:teaser}
\end{figure*}

\begin{abstract}
Training large language models (LLMs) as interactive agents for controlling graphical user interfaces (GUIs) presents a unique challenge to optimize long-horizon action sequences with multimodal feedback from complex environments.
While recent works have advanced multi-turn reinforcement learning (RL) for reasoning and tool-using capabilities in LLMs, their application to GUI-based agents remains relatively underexplored due to the difficulty of sparse rewards, delayed feedback, and high rollout costs.
In this paper, we investigate \textbf{end-to-end policy optimization} for vision-language-based GUI agents with the aim of improving performance on complex, long-horizon computer tasks.
We propose \textbf{A}gentic \textbf{R}eplay \textbf{P}olicy \textbf{O}ptimization~(\textbf{ARPO}), an end-to-end RL approach that augments Group Relative Policy Optimization (GRPO) with a replay buffer to reuse the successful experience across training iterations.
To further stabilize the training process, we propose a task selection strategy that filters tasks based on baseline agent performance, allowing the agent to focus on learning from informative interactions.
Additionally, we compare ARPO with offline preference optimization approaches, highlighting the advantages of policy-based methods in GUI environments. Experiments on the OSWorld benchmark demonstrate that ARPO achieves competitive results, establishing a new performance baseline for LLM-based GUI agents trained via reinforcement learning.
Our findings underscore the effectiveness of reinforcement learning for training multi-turn, vision-language GUI agents capable of managing complex real-world UI interactions. Codes and models:\url{https://github.com/dvlab-research/ARPO.git}.
\end{abstract}

%%%%%%%%% MAIN PAPER %%%%%%%%%
%%%%%%%%% INTRODUCTION %%%%%%%%%
\section{Introduction}

Among various agent types, GUI agents that interact with the computer screen through vision-based perception and action have been of long-standing interest~\cite{cheng2024seeclick, qin2025ui, gou2024navigating}. Most prior work relies on supervised fine-tuning (SFT) on large-scale trajectory data. These agents are typically trained through SFT  on large-scale trajectory datasets, where the model learns to imitate human behavior by predicting the next action based on the current screenshot and interaction history. However, these agents lack the ability to self-correct and suffer from error accumulation in the working trajectory.

To address these limitations, we explore reinforcement learning (RL) in GUI agent training. In contrast to single-turn RL or static reward optimization, we adopt Group Relative Policy Optimization (GRPO)~\cite{guo2025deepseek}, a recent variant of PPO~\cite{schulman2017proximal} that eliminates the need for a value function and estimates token-level advantages from group-wise reward normalization. GRPO has demonstrated promising results in mathematical reasoning~\cite{shao2402deepseekmath} and tool-use agent~\cite{qian2025toolrl}. It is a natural fit for training vision-language agents due to its ability to handle long sequences and multiple modalities with improved efficiency.

This paper tackles the challenge of end-to-end policy optimization for GUI agents, with a particular focus on multi-turn, multi-modal agent design, see Fig.~\ref{fig:teaser_a}. Our goal is to maximize the rule-based reward from the environment over entire trajectories using GRPO. However, GUI environments typically offer sparse and delayed reward signals: agents receive feedback only upon task completion, and many complex tasks may yield no reward at all during early training phases. Moreover, the cost of rollouts in real desktop environments is non-trivial. GUI interaction involves operating system-level delays, which significantly slow down the data collection process. To overcome these obstacles, we develop a scalable distributed rollout system that enables parallel interaction with real desktop environments, such as OSWorld~\cite{xie2024osworld}. By batching inference across environments, we reduce latency and make efficient use of GPU resources, thus facilitating rollout collection at scale.

% To address the sparse-reward nature of GUI tasks and high rollout cost, we introduce a scalable distributed rollout system for parallel interaction with real desktop environments (e.g., OSWorld~\cite{xie2024osworld}). We further propose an experience replay buffer that stores successful trajectories and ensures that each GRPO training group contains reward diversity. This improves learning stability and allows effective gradient updates even in low-success-rate regimes.
To further enhance training stability and sample efficiency, we introduce a task selection strategy that filters for those capable of producing successful rollouts under baseline agents. This curated subset enhances signal quality during early training and accelerates convergence. We further introduce an experience replay buffer tailored to GUI agent learning. This buffer stores successful trajectories on a per-task basis and dynamically injects them into GRPO training groups when all sampled rollouts in a group fail. The inclusion of at least one high-reward trajectory within each group ensures meaningful reward variance, which is critical for computing token-level advantages. 

We conduct extensive evaluations on the OSWorld benchmark and observe that reinforcement learning effectively improves agent performance. We also find an interesting fact that RL training delivers strong gains on in-domain tasks, but hardly benefits out-of-domain agentic tasks. 

Our contributions are summarized as follows:

\begin{itemize}[leftmargin=10mm, itemsep=4pt, parsep=0pt]
    \item We propose an end-to-end policy optimization approach for training a GUI agent in challenging multi-turn, multi-modal environments using GRPO.
    \item We demonstrate that careful selection of training tasks is critical for maintaining reward diversity and ensuring stable policy optimization.
    \item We propose an experience replay buffer that retains successful trajectories, enhancing sample efficiency and stabilizing training in sparse-reward settings.
    \item We find that reinforcement learning substantially improves agent performance on in-domain tasks, while offering moderate generalization improvements to out-of-domain agentic tasks.
\end{itemize}

%%%%%%%%% RELATED WORKS %%%%%%%%%
\section{Related Works}

\paragraph{GUI Agents.} 
Recent advances in multimodal models have led to significant progress in GUI and web-based automation. SeeClick~\cite{cheng2024seeclick} and ScreenAgent~\cite{niu2024screenagent} utilize large vision-language models (VLMs) with visual input processing to perform interactive tasks on user interfaces. Building on this, OmniAct~\cite{kapoor2024omniact} introduces a benchmark that focuses on generating executable actions from visually grounded natural language instructions. CogAgent~\cite{hong2024cogagent} and UI-Tars~\cite{qin2025ui} extend pretraining with large-scale web and desktop interface data, enhancing screen understanding and agent behavior. GUI-R1~\cite{xia2025gui} explores reinforcement learning to improve UI grounding in VLM-based agents. However, directly optimizing policy models for GUI agents in an end-to-end policy optimization way remains unexplored in current research.

\paragraph{Reinforcement Learning for Agents.}
 % Rule-based reinforcement fine-tuning, exemplified by OpenAI o1~\cite{jaech2024openai} and DeepSeek-R1~\cite{guo2025deepseek}, has demonstrated strong performance in mathematical reasoning~\cite{shao2402deepseekmath}, code-generation~\cite{liu2025code}, and multi-modal reasoning~\cite{huang2025vision, liu2025seg}. ToolRL~\cite{qian2025toolrl} takes a step further to utilize RL for tool-use LLM agent training. Sweet-RL~\cite{zhou2025sweet} proposes a multi-turn DPO to boost performance of the LLM agent. RAGEN~\cite{wang2025ragen} proposes a multi-turn RL training approach in rule-based live environment. 
Rule-based reinforcement learning (RL) has proven effective in fine-tuning large language models (LLMs) across a range of domains. OpenAI's o1~\cite{jaech2024openai} and DeepSeek-R1~\cite{guo2025deepseek} demonstrate strong performance in tasks such as mathematical reasoning~\cite{shao2402deepseekmath}, code generation~\cite{liu2025code}, and multi-modal inference~\cite{huang2025vision, liu2025seg} through structured reward signals. ToolRL~\cite{qian2025toolrl} extends this paradigm by introducing RL-based training for LLM agents that interact with external tools. Sweet-RL~\cite{zhou2025sweet} introduces a multi-turn DPO framework to enhance long-horizon language agent behaviors. RAGEN~\cite{wang2025ragen} further advances multi-turn RL by applying it in live, rule-based environments for self-evolving agent training.

Despite these advancements, most existing work focuses on symbolic tasks or static tool use. Applying reinforcement learning to vision-language agents operating in dynamic, multimodal GUI environments remains a challenging task. In particular, this work aims to leverage rule-based rewards from live desktop environments for end-to-end policy optimization in multi-turn GUI agents.

\section{Method}
In this section, we first provide a brief introduction to the preliminaries of Group Relative Policy Optimization~(GRPO). Then, we describe the architecture and training procedure of our GUI agent. The agent builds upon vision-language models (VLMs), enhanced with longer context windows and longer image, action chains. These modifications are essential for training complex GUI tasks with end-to-end reinforcement learning algorithms like GRPO.

\subsection{Preliminary}
Group Relative Policy Optimization (GRPO)~\cite{shao2402deepseekmath} is a reinforcement learning algorithm designed to optimize language models efficiently without requiring an explicit value function or critic. GRPO modifies the standard Proximal Policy Optimization (PPO) objective by computing token-level advantages based on group-normalized rewards, making it particularly suitable for LLMs.

Given a batch of $G$ responses $\{o_i\}_{i=1}^G$ from a query $q$, each consisting of a sequence of tokens $o_i = (o_i(1), ..., o_i(T))$, the GRPO objective is defined as:
\begin{equation}
J_{\text{GRPO}}(\theta) = \frac{1}{G} \sum_{i=1}^{G} \frac{1}{|o_i|} \sum_{t=1}^{|o_i|}
\min \left[
\frac{\pi_\theta(o_i(t) | o_{i,<t})}{\pi_{\text{old}}(o_i(t) | o_{i,<t})} \hat{A}_{i,t},\,
\text{clip} \left(
\frac{\pi_\theta(o_i(t) | o_{i,<t})}{\pi_{\text{old}}(o_i(t) | o_{i,<t})}, 1 - \varepsilon, 1 + \varepsilon
\right) \hat{A}_{i,t}
\right], \notag
\end{equation}
where $\hat{A}_{i,t}$ is the group-normalized advantage for token $t$ in response $o_i$, computed as:
\[
\hat{A}_{i,t} = \frac{r_i - \mu}{\sigma}, \quad \text{with } r_i \text{ the total reward of } o_i,
\]
and $\mu$, $\sigma$, the mean and standard deviation of rewards in the group.
\begin{figure}[t!]
\centering
    \includegraphics[width=0.95\linewidth]{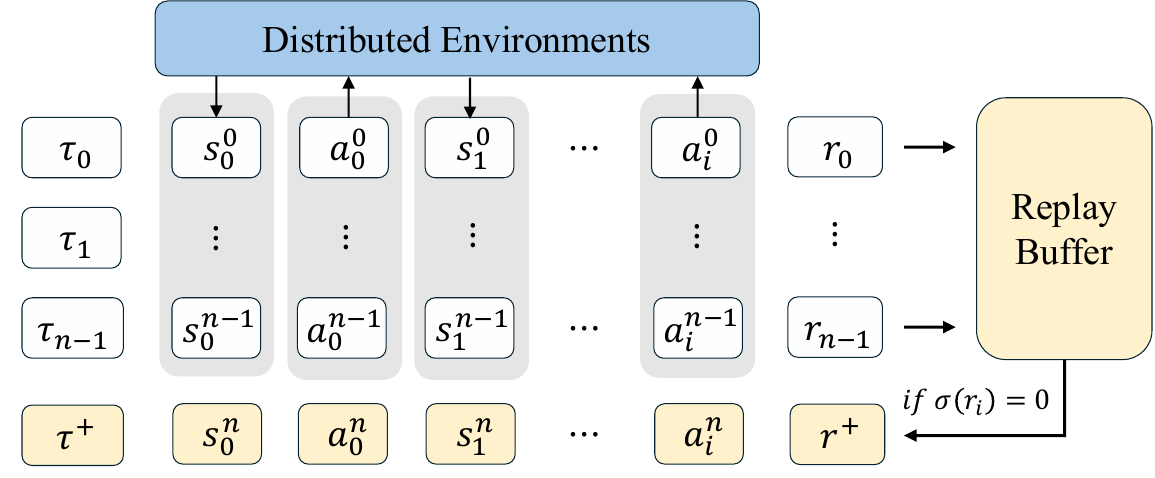}
    \caption{Illustration of the reinforcement learning procedure for our multi-turn GUI agent. For a single task, we use $n$ parallel environments and perform rollouts to collect trajectories and rewards $\{\tau_i, r_i\}_{i=0, 1, \cdots, n-1}$ in the environments. If all the rewards are zero, we fetch a positive trajectory $\tau^+$ from the replay buffer to avoid gradient vanishing.
    }
\label{fig:framework}
\end{figure}

\subsection{Multi-turn GUI Agent}
Unlike single-turn reinforcement learning, GUI agents are required to perform multi-turn reasoning and decision-making, interacting with dynamic environments that provide visual feedback. We adopt a Markov Decision Process (MDP) framework, where each agent trajectory comprises a sequence of screenshot observations $s_t$, mouse and keyboard actions $a_t$, and a scalar reward $r$ in the end of the trajectory. The agent policy $\pi_\theta$ is optimized to maximize the rewards:
\begin{equation}
    \tau = \{s_t, a_t\}_{t=0, 1, \cdots, T-1}\text{,\ \  where } a_t \sim P_\theta\left(\left\{s_i, a_i\right\}_{i < t}\right).
\end{equation}
Our GUI agent is built upon the UI-Tars~\cite{qin2025ui} framework and the Qwen2.5-VL architecture~\cite{bai2025qwen2}. To predict the next action $a_t$ the model tokenizes the entire history of screenshots and corresponding actions into the input context of the VLM model. 

Our design results in a VLM model with at most $15$ images input and $64K$ model context length to correctly process an entire GUI trajectory with 1080P resolution. Unlike prior short-context GUI agents~\cite{cheng2024seeclick, gou2024navigating, xu2024aguvis}, which truncate the trajectory and only process the most recent one or two screenshots, our approach leverages the full trajectory history, enabling the model to reason over long-term dependencies and optimize performance across the entire interaction sequence.

% To enhance the performance of VLM agent with chain-of-thought~\cite{wei2022chain} skill, each action $a_t$ consists of a thinking and a solution response. Our GUI Agent follows the action space definition of UI-Tars, which contains left click, right click, scrolling, typing and hotkeys. There are additional actions like WAIT, FINISH, FAIL, CALL\_USER to control the workflow of the agent.
\begin{tcolorbox}[colback=lightblue, colframe=blue!50!black, title=CoT for GUI Agents]
To enhance the reasoning capabilities of VLM agents, we integrate the \textbf{Chain-of-Thought (CoT)} prompting technique~\cite{wei2022chain} into our action generation process. Each action $a_t$ is composed of:
\begin{itemize}
    \item A \textbf{thinking part}, which represents the agent’s internal reasoning.
    \item A \textbf{solution part}, which executes the resulting action.
\end{itemize}
This design allows the agent to perform more accurate and interpretable decision-making.
\end{tcolorbox}

\begin{tcolorbox}[colback=lightgray, colframe=black, title=Action Space Definition]
Our GUI agent adopts the action space defined in UI-Tars~\cite{qin2025ui}, including the following primitive operations:
\begin{itemize}
    \item \texttt{LEFT\_CLICK}, \texttt{RIGHT\_CLICK}, \texttt{SCROLL}
    \item \texttt{TYPE\_TEXT}, \texttt{PRESS\_HOTKEY}
\end{itemize}

In addition to these, several meta-actions are used to manage the agent's workflow:
\begin{itemize}
    \item \texttt{WAIT}: Pause and observe the environment.
    \item \texttt{FINISH}: Successfully complete the task.
    \item \texttt{FAIL}: Indicate task failure.
    \item \texttt{CALL\_USER}: Request human intervention.
\end{itemize}
\end{tcolorbox}

\subsection{Distributed Trajectory Rollout}
Training GUI agents through reinforcement learning requires scalable and efficient trajectory collection across rich, interactive desktop environments. To meet this need, we design a distributed trajectory rollout strategy tailored for parallel interaction with live environments, such as OSWorld~\cite{xie2024osworld}. 

We establish a set of rollout workers. Each worker consists of an interactive environment paired with a GUI agent that maintains a history of screenshots and corresponding actions, denoted as $(s_t, a_t)$. Each rollout worker continuously captures screenshots of the current GUI environment and transmits them to a centralized language model inference server powered by VLLM~\cite{kwon2023efficient}. The policy model processes these batched visual observations in parallel, predicting the next action for all environments simultaneously.

Unlike math~\cite{shao2402deepseekmath} or tool-use~\cite{qian2025toolrl} environments, interaction with live GUIs like OSWorld incurs non-trivial latency due to OS-level delays. Parallel trajectory rollout allows for efficient utilization of GPU resources on the inference server and minimizes the per-step decision latency.

\subsection{End-to-End Policy Optimization with GRPO}
\label{subsec:GPRO}
We adopt GRPO~\cite{shao2402deepseekmath} as our reinforcement learning algorithm to train vision-based GUI agents. GRPO eliminates the need for a value function by leveraging group-wise reward normalization to compute token-level advantages. This property makes it well-suited for training VLM agents with multiple image inputs and extended context length.

\paragraph{Reward Design.}
To effectively guide policy optimization, we design a structured reward function incorporating both task-level success and action format correctness.

\begin{itemize}[leftmargin=5mm]
    \item \textbf{Trajectory Reward:} For each task, we have a scalar trajectory-level reward $r_t$, based on task completion. A reward of $r_t=1$ is assigned if the agent successfully completes the task as defined by the OSWorld~\cite{xie2024osworld}, and $r_i=0$ otherwise. This binary reward provides a high-level training signal to encourage successful multi-turn planning and execution.
    \item \textbf{Action Format Reward:} During rollout, each response from the VLM agent is parsed into discrete actions. If a response fails to conform to the required action schema and cannot be parsed, we assign a penalty of $r_f=-1$. This encourages the model to generate syntactically valid and executable actions.
\end{itemize}

\paragraph{Training Objective.} We treat GUI interaction as a multi-turn MDP, where the agent observes a sequence of screenshots \( s_t \) and generates actions \( a_t \) to complete a task instruction \( x \in \mathcal{D} \). The trajectory \( \tau = (s_0, a_0, \ldots, a_n)\) is encoded by a VLM agent with extended context, enabling long-horizon reasoning over multiple steps and observations. Our training objective is to maximize the expected reward over tasks and trajectories:

\begin{equation}
    \max_{\theta} \; \mathbb{E}_{x \sim \mathcal{D}, \, \tau \sim \pi_\theta} \left[ r_t\left(x, \tau\right) + r_f\left(x, \tau\right) \right].
\end{equation}

We optimize this objective using GRPO, which estimates token-level advantages via group-normalized trajectory rewards, allowing efficient and scalable training without a value function.

\paragraph{Valuable Tasks Selection for GRPO.} Despite recent progress in the variants of GRPO~\cite{yu2025dapo}, the task of training GUI agents remains difficult, particularly due to the sparse reward signals associated with complex desktop environments like OSWorld~\cite{xie2024osworld}. Many tasks in this benchmark are not reliably solvable by current state-of-the-art agents~\cite{qin2025ui, gou2024navigating}, even when given multiple attempts. As a result, these tasks generate limited feedback during rollouts, which can hinder effective policy optimization training for GRPO.

To improve the sampling efficiency, we introduce a task filtering procedure to identify a subset of ``valuable'' tasks, those capable of producing successful trajectories under a baseline agent. Specifically, we evaluate each task in OSWorld using the UI-Tars-1.5 model, performing 16 rollouts per task. A task is retained in the GRPO training set if the agent completes it successfully in at least one of these attempts. This method yields a curated set of 128 tasks that are more amenable to early-stage learning, allowing the policy optimization to benefit from informative reward signals. 

\subsection{Experience Replay Buffer}
Dynamic Sampling~\cite{yu2025dapo} has been proposed to improve the sample efficiency of GRPO by removing training groups in which all rewards are uniform. In such cases, the computed token-level advantages are zero across the group, resulting in vanishing gradients and slowed convergence. However, this strategy becomes less effective in GUI interaction settings due to two primary challenges: the high cost of obtaining trajectories and the infrequency of successful rollouts.

Unlike mathematical reasoning tasks, which typically follow well-defined logical chains, GUI-based tasks sometimes require a certain amount of exploratory interactions with the environment, resulting in sparse reward signals. Therefore, successful trajectories are rare but especially informative. Preserving and reusing them is critical for the training progress.

To address this, we introduce an experience replay buffer that caches successful trajectories on a per-task basis. During training, if an entire GRPO training group consists of only failed trajectories (\textit{i.e.}, all with zero reward), we randomly replace one of them with a previously stored successful trajectory from the buffer for the corresponding task. This guarantees that, as long as the agent has successfully completed a task once, its training group in the later training process will include at least one rollout with a non-zero reward signal, as illustrated in Fig.~\ref{fig:framework}. The buffer is updated dynamically during rollout. To prevent the stored samples from diverging too significantly from the current policy, we impose a fixed-size limit on the buffer and evict the oldest entries when full.
% Select a subset of successful tasks for GRPO

\section{Experiments}

\subsection{Implementation Details}
\paragraph{Training Details.} We use the 7B UI-Tars-1.5 model~\cite{qin2025ui} as the base and conduct training using the VERL framework~\cite{sheng2024hybridflow}. For trajectory rollout, we set up 256 parallel virtual environments and the rollout number for each task is 8. A total of 128 tasks are sampled from the OSWorld benchmark, according to the strategy described in Sec.~\ref{subsec:GPRO}, and training is performed over 15 epochs. Rollouts are conducted with a batch size of 32 and a temperature of 1.0 to encourage exploration. For policy optimization, we use the AdamW optimizer~\cite{loshchilov2017decoupled} with a learning rate of $1 \times 10^{-6}$ and a mini-batch size of 8 per device. The gradient accumulation number is 4. Following DAPO~\cite{yu2025dapo}, we set the clipping parameters to $\epsilon_{\text{low}} = 0.2$ and $\epsilon_{\text{high}} = 0.3$ to balance exploration and exploitation. During evaluation, the temperature is lowered to 0.6 for more stable performance. We remove the KL divergence loss to remove the need for the reference model.

\paragraph{Datasets and Benchmarks.} We evaluate our method on the OSWorld benchmark~\cite{xie2024osworld}, a recently proposed real-computer environment designed for evaluating multimodal agents on open-ended GUI tasks. OSWorld contains 369 tasks across diverse domains such as office productivity, web browsing, system management, and multi-app workflows. Each task is executed within virtual machines using real applications and evaluated via execution-based scripts. The benchmark supports full GUI interaction with mouse and keyboard actions, enabling rigorous assessment of multi-turn vision-based agents in realistic desktop environments.

\paragraph{Evaluation Metrics.} We follow the standard rule-based evaluation protocol defined in OSWorld~\cite{xie2024osworld}. Each agent trajectory receives a scalar reward between 0 and 1.0 from the environment. We notice that previous works~\cite{qin2025ui} replace the last action with a FAIL action when the maximum step number is reached in a rollout. While this approach may prevent unstable or endlessly running behaviors during evaluation, it will hack the rewards of the real impossible tasks defined in the benchmark. To provide a more accurate assessment of agent capabilities for RL, we introduce a stricter evaluation protocol that prohibits final action replacement, denoted as \textbf{OSWorld Hard}.
%To clearly evaluate the models, we split all the tasks in OSWorld into two categories  

\subsection{Experimental Results}
We evaluate the performance of our ARPO method on the OSWorld benchmark~\cite{xie2024osworld}, comparing it against several recent GUI agents. As shown in Table~\ref{tab:all}, our approach achieves the highest success rates across both evaluation settings. Specifically, applying ARPO to the UI-Tars-1.5 base model results in a success rate of $29.9\%$ on the standard OSWorld setting and $23.8\%$ on the stricter OSWorld Hard variant—improving upon the original UI-Tars-1.5 model by 6.4\% and 5.6\%, respectively. These results highlight the effectiveness of reinforcement learning with GRPO and structured experience replay in enhancing multi-turn GUI decision-making. Additionally, ARPO shows consistent gains across earlier model versions; for example, UI-Tars-7B-DPO improves from $15.6\%$ to $20.4\%$ with ARPO. All the models are tested with a maximum step number limit of 15 for a single trajectory.

\begin{table}[t]
\centering
\caption{OSWorld evaluation performance for GUI Agents. All models are evaluated at a maximum execution length of 15. We provide numerical results for two metrics: \textbf{OSWorld} and \textbf{OSWorld Hard}}
\begin{adjustbox}{max width=\textwidth}
% \begin{tabular}{l*{4}{S[table-format=2.1, detect-weight=true, detect-inline-weight=math]}}
\begin{tabular}{y{105}x{60}y{70}y{70}}
\toprule
\textbf{Model} & \textbf{GPT-4o} & \textbf{OSWorld} & \textbf{OSWorld Hard} \\
\midrule
Aria-UI~\cite{yang2024aria} & \checkmark  & 15.2\% & \text{-} \\
Aguvis-7B~\cite{xu2024aguvis} & \checkmark & 14.8\%  & \text{-} \\
Aguvis-72B~\cite{xu2024aguvis} & \checkmark & 17.0\%  & \text{-} \\
OS-Atlas-7B~\cite{wu2024atlas} &\checkmark & 14.6\%  & \text{-} \\
\midrule
UI-Tars-7B-DPO &  & 15.6\%   & 11.3\%  \\
UI-Tars-7B-DPO + GRPO &  & 18.3\% \cgaphlb{+}{2.7\%} & 16.4\% \cgaphlb{+}{5.1\%}  \\
\cellcolor[HTML]{E7F0F9} UI-Tars-7B-DPO + ARPO &  \cellcolor[HTML]{E7F0F9} & \cellcolor[HTML]{E7F0F9} \textbf{20.4\%}\cgaphl{+}{4.8\%} & \cellcolor[HTML]{E7F0F9} \textbf{18.0\%}\cgaphl{+}{6.7\%} \\
UI-Tars-7B-1.5        &  & 23.5\% & 18.2\% \\
UI-Tars-7B-1.5 + GPRO &  & 26.0\% \cgaphlb{+}{2.5\%} & 20.9\% \cgaphlb{+}{2.7\%} \\
\cellcolor[HTML]{E7F0F9} UI-Tars-7B-1.5 + ARPO & \cellcolor[HTML]{E7F0F9}  & \cellcolor[HTML]{E7F0F9} \textbf{29.9\%}\cgaphl{+}{6.4\%} & \cellcolor[HTML]{E7F0F9} \textbf{23.8\%}\cgaphl{+}{5.6\%}\\
\bottomrule
\end{tabular}
\end{adjustbox}
\label{tab:all}
\end{table}
\subsection{Ablation on the Replay Buffer}

\begin{figure}[t!]
\centering
\begin{minipage}[t]{0.45\linewidth}
    \centering
    \includegraphics[width=\linewidth]{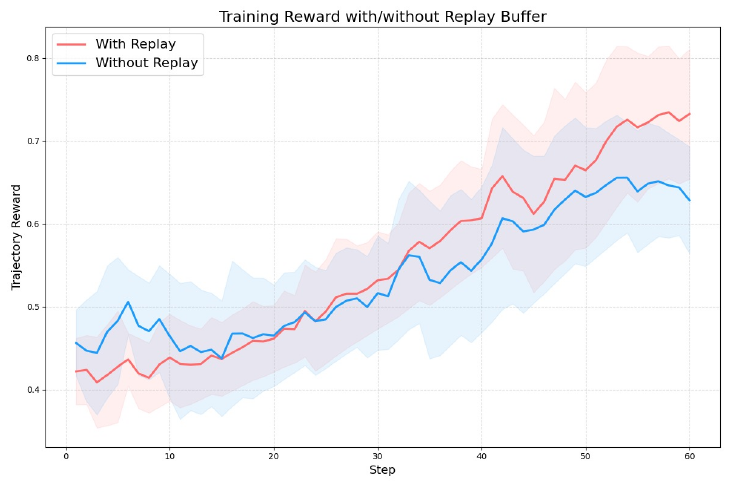}
    \caption{Ablation study of the replay buffer.}
    \label{fig:ablation_replay}
\end{minipage}
\hfill
\begin{minipage}[t]{0.52\linewidth}
    \centering
    \includegraphics[width=\linewidth]{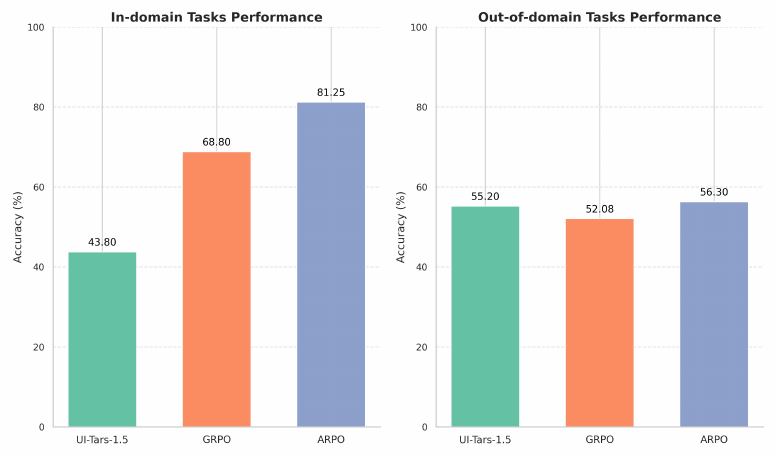}
    \caption{Ablation study for GRPO and ARPO in-domain and out-of-domain RL training tasks.}
    \label{fig:ablation_replay2}
\end{minipage}
\end{figure}

We evaluate the impact of the experience replay buffer by comparing training trajectories with and without its use, as shown in Fig.~\ref{fig:ablation_replay}. The model equipped with the replay buffer begins to outperform the baseline around Step 30 and maintains a consistent advantage throughout the remainder of training. This improvement is attributed to the buffer’s ability to retain successful trajectories. This gain is largely due to the buffer's ability to retain high-reward trajectories from earlier stages, which serve as strong learning signals in later updates. By maintaining reward diversity within GRPO groups and non-zero advantages, the replay buffer supports more stable optimization and accelerates convergence. By the end of training, the model with replay achieves a higher average trajectory reward (0.75 vs. 0.65), demonstrating that leveraging past successes substantially improves both sample efficiency and overall policy performance in sparse-reward GUI environments.

The benefits of the replay buffer extend beyond reward curves. As shown in Fig.~\ref{fig:ablation_replay2}, the in-domain task success rate climbs from 68.8\% with GRPO to 81.25\% with ARPO, a 12.5\% absolute improvement. This substantial gain highlights the replay buffer’s critical role in enhancing policy generalization and downstream performance. 

\subsection{Does RL training generalize well to OOD GUI agent tasks?}
To assess the generalization ability of RL training, we evaluate model performance on both in-domain and out-of-domain (OOD) tasks. Specifically, we select 32 tasks from the training task set for reinforcement learning, using the remaining 96 as OOD tasks. As shown in Fig.~\ref{fig:ablation_replay2}, reinforcement learning substantially improves in-domain accuracy: GRPO achieves 68.8\% and ARPO reaches 81.25\%, compared to 43.8\% for the base UI-Tars-1.5 model. However, on OOD tasks, gains are more modest. UI-Tars-1.5 achieves 55.2\%, while GRPO slightly underperforms at 52.08\%. ARPO, however, recovers generalization capability, scoring 56.3\%, slightly above the base model, indicating that structured trajectory grouping and replay mitigate overfitting. Overall, while reinforcement learning effectively improves the in-domain success rate of VLM agents, strong generalization still depends on broader task diversity, carefully designed reward signals, and larger-scale training compute.

\subsection{Valuable Task Selection for GRPO Training}
\begin{figure}[t!]
\centering
\begin{subfigure}[t]{0.49\linewidth}
    \centering
    \includegraphics[width=\linewidth]{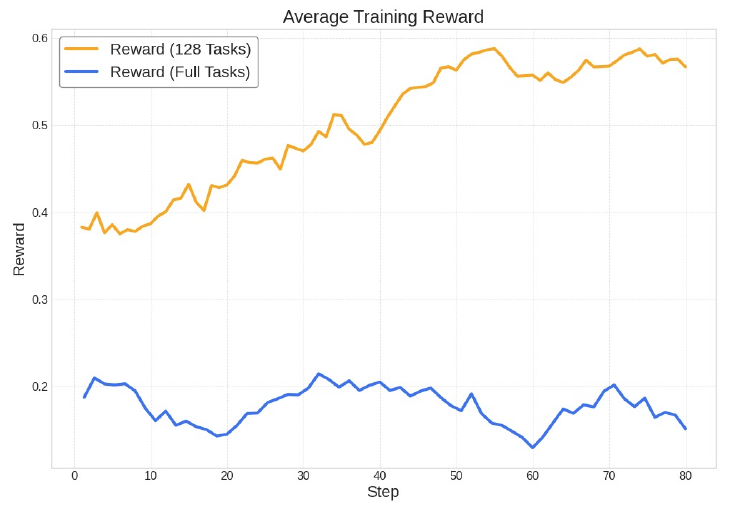}
    \caption{Training rewards.}
    \label{fig:reward_full}
\end{subfigure}
\hfill
\begin{subfigure}[t]{0.49\linewidth}
    \centering
    \includegraphics[width=\linewidth]{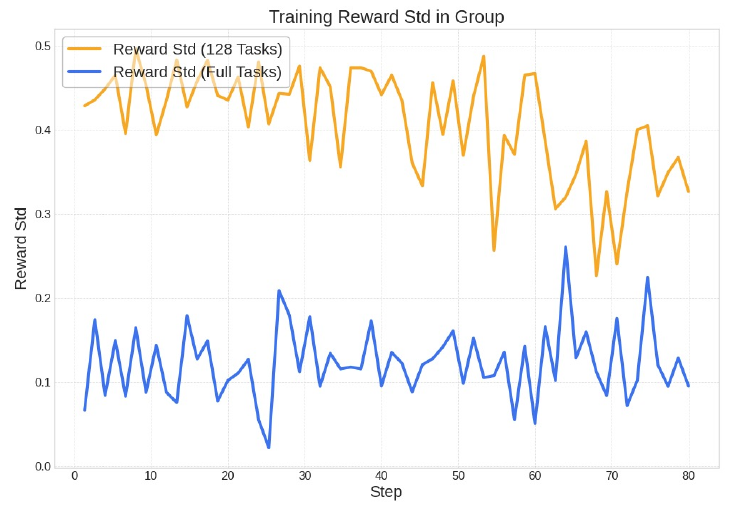}
    \caption{Reward variance during RL training.}
    \label{fig:reward_std_full}
\end{subfigure}
\caption{Training performance comparison for RL training with selected subset and full set.}
\label{fig:rl_training}
\end{figure}

As outlined in Sec.~\ref{subsec:GPRO},  we adopt a task selection strategy for GRPO training by filtering out tasks that consistently fail to provide meaningful reward signals. To evaluate the impact of this approach, we conduct an ablation study comparing GRPO performance when trained on a curated subset of 128 valuable tasks versus the full task set. As illustrated in Fig.~\ref{fig:reward_full}, training on the selected subset leads to significantly higher average trajectory rewards and faster convergence speed from the early stages of training. 

Fig.~\ref{fig:reward_std_full} shows that the standard deviation of rewards within GRPO groups is consistently higher when training on the curated task set. This increased variance is critical for GRPO, which relies on within-group reward diversity to compute token-level advantages. In contrast, training on the full task set results in flatter reward distributions with reduced variance. 

% \begin{figure}[t!]
% \centering
% \begin{minipage}[t]{0.49\linewidth}
%     \centering
%     \includegraphics[width=\linewidth]{figs/smoothed.pdf}
%     \caption{Rewards during RL training for selected training subset and full set.}
%     \label{fig:reward_full}
% \end{minipage}
% \hfill
% \begin{minipage}[t]{0.49\linewidth}
%     \centering
%     \includegraphics[width=\linewidth]{figs/reward_std.pdf}
%     \caption{Reward Variance during RL training.}
%     \label{fig:reward_std_full}
% \end{minipage}
% \end{figure}

\subsection{Comparison with Offline Preference Optimization}
\begin{figure}
\centering
\begin{minipage}[t]{0.49\linewidth}
    \centering
    \includegraphics[width=\linewidth]{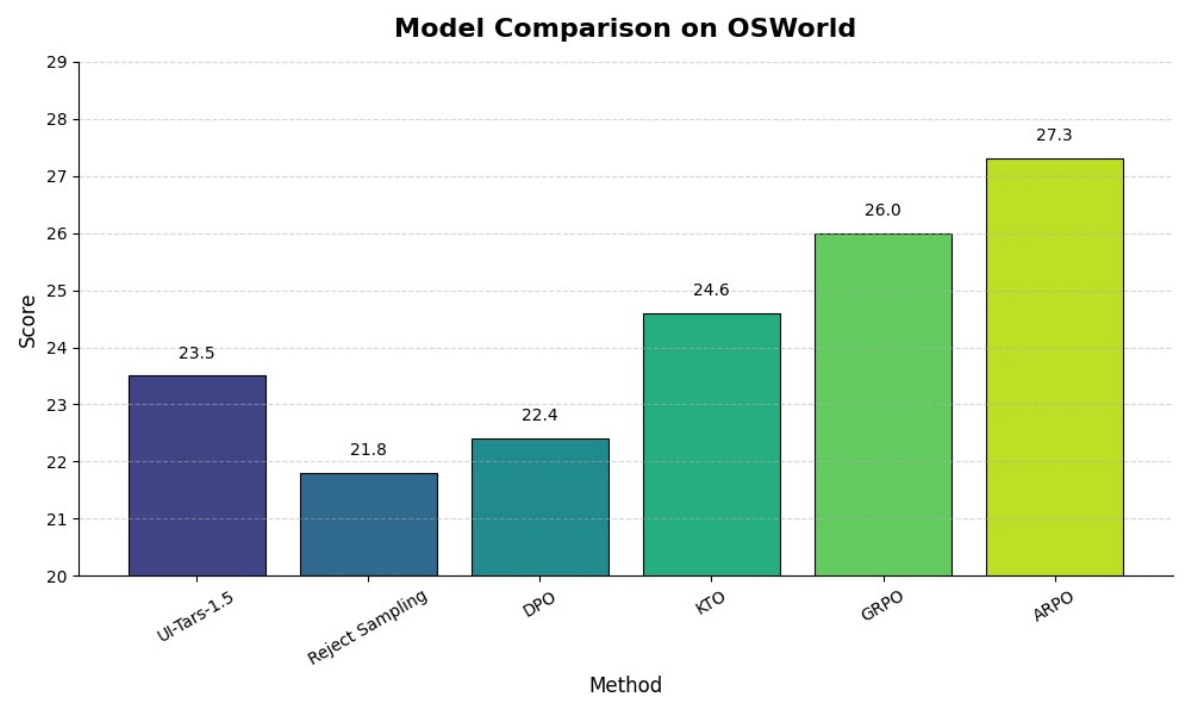}
    \caption{Comparisons with offline preference optimization methods.}
    \label{fig:preference}
\end{minipage}
\hfill
\begin{minipage}[t]{0.49\linewidth}
    \centering
    \includegraphics[width=\linewidth]{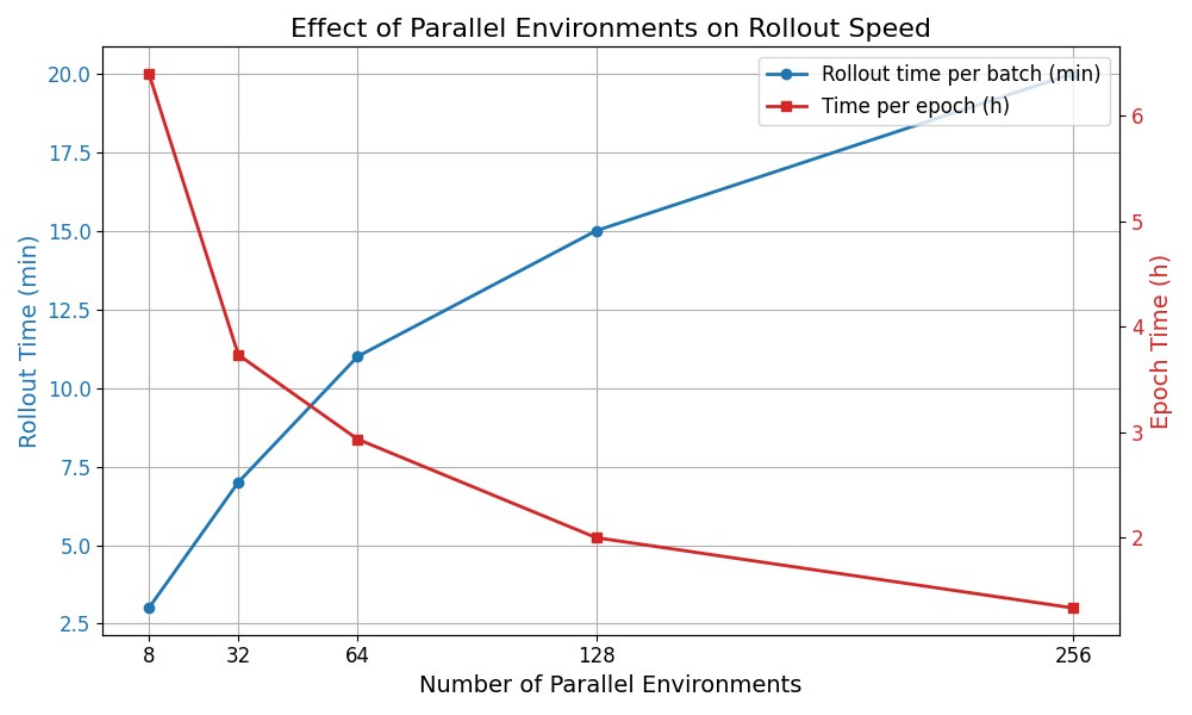}
    \caption{Rollout Efficiency.}
    \label{fig:speed}
\end{minipage}

\end{figure}

In Fig.~\ref{fig:preference}, we compare the performance between GRPO and offline preference optimization algorithms. For a fair comparison, all methods are trained on the same task set with an equal number of rollouts. We compare GRPO with reject sampling, DPO~\cite{rafailov2023direct}, and KTO~\cite{ethayarajh2024kto}. For reject sampling, we take only the positive trajectory for SFT training. For DPO, we randomly sample a positive and a negative trajectory per task to create paired training data. For KTO, we threshold the scalar rewards at 0.5 to generate binary labels for training.

ARPO achieves the highest score (27.3\%), followed by GRPO (26.0\%), both outperforming preference-based methods by a significant margin. Among the baselines, KTO performs best (24.6\%), while DPO and Reject Sampling lag behind at 22.4\% and 21.8\% , respectively. These results suggest that direct trajectory-level optimization with rule-based rewards provides stronger learning signals than offline preference modeling. The added experience replay in ARPO further enhances stability and sample efficiency in sparse-reward GUI settings.

\subsection{Rollout Efficiency Analysis}
Fig.~\ref{fig:speed} shows that increasing the number of parallel environments significantly improves training efficiency. We show the rollout time for a single batch of trajectories (in minutes) and the total time to sample all trajectories in an epoch (in hours), respectively. As the batch size increases, although the rollout time per batch grows from ~3 minutes (8 environments) to ~19 minutes (256 environments), the total time per epoch drops sharply from over 6 hours to just around 1.2 hours.

This speedup stems from two main factors: (1) Larger batches allow the VLLM server to perform more efficient GPU inference, and (2) OS-level delays in GUI environments are overlapped for all parallel environments. As a result, scaling to 256 environments enables high-throughput rollouts, making RL training in real desktop settings much more practical.

\subsection{Qualitative Analysis: Self-Correction Behavior in GUI Agent}
\begin{figure}
    \centering
    \includegraphics[width=\linewidth]{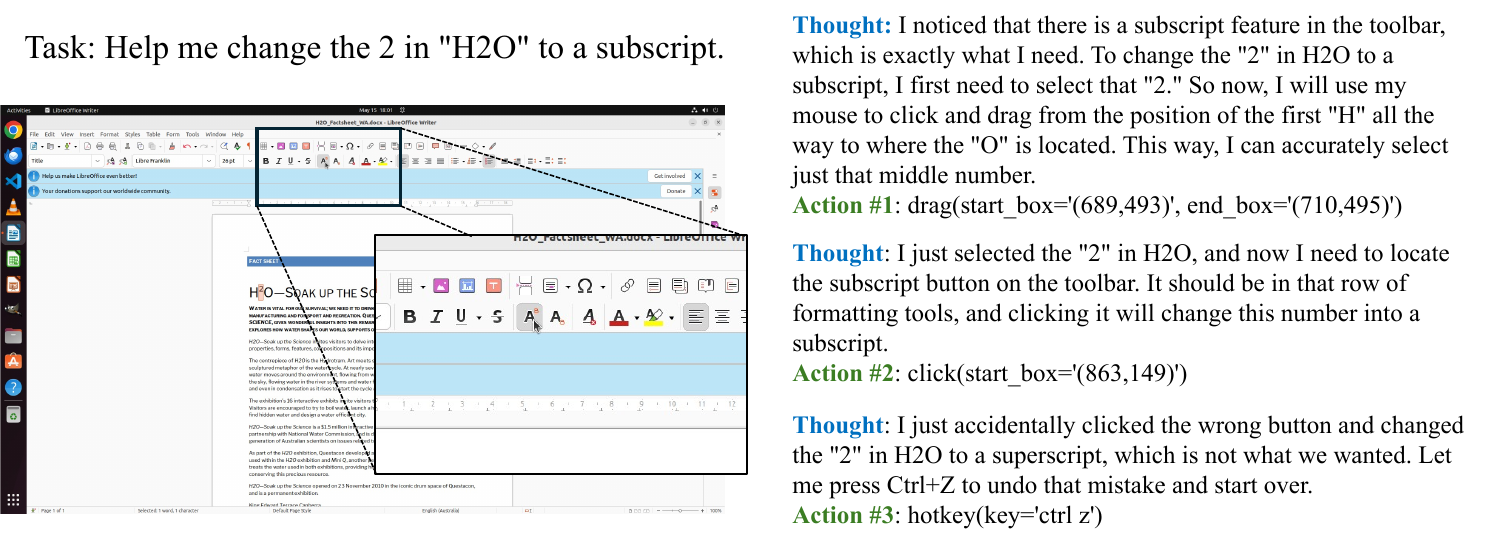}
    \caption{Trajectory showing the ARPO-trained agent correcting a misclick error.}
    \label{fig:visual}
\end{figure}

Figure~\ref{fig:visual} illustrates a trajectory where the ARPO-trained agent demonstrates self-corrective behavior. Initially, it selects the superscript button instead of the subscript button. It realizes the mistake by observing the current screen and decides to use the Ctrl-Z hotkey to revert the previous operation. Notably, the success rate for the specific before and after ARPO are 25\% vs. 62.5\%.

%%%%%%%%% CONCLUSION %%%%%%%%%
\section{Conclusion}

In this work, we present a reinforcement learning approach for training GUI agents using vision-language models enhanced with longer input context and multi-turn, multi-modal screenshot processing. By introducing ARPO, a variant of GRPO tailored for GUI agents, we demonstrate that rule-based reward signals can effectively guide end-to-end policy optimization in complex GUI environments. Our experiments show that careful task selection significantly improves learning stability and reward variance.

This study highlights the potential of combining multimodal understanding with reinforcement learning to enable more adaptive and capable GUI agents. Future directions include expanding the task set to cover a broader range of real-world applications, extending the context length of agents further to support more sophisticated trial-and-error behaviors, and investigating the use of learned reward models to autonomously evaluate trajectories, reducing reliance on manually crafted reward functions.

% \input{others/nerips_supp_arxiv}

%%%%%%%%% REFERENCE %%%%%%%%%
% \newpage
{
  \small
  \bibliographystyle{plain}
  \bibliography{neurips_2023}
}
%%%%%%%%%%%%%%%%%%%%%%%%%%%%%%%%%%%%%%%%%%%%%%%%%%%%%%%%%%%%

\end{document}

%% file: others/shortcuts.tex
\usepackage{times}
\usepackage{epsfig}
\usepackage{graphicx}
\usepackage{float}
\usepackage{wrapfig}
\usepackage{amsmath,amssymb,amsthm}
\usepackage{algorithm,algorithmicx,algpseudocode}
\usepackage{bm,xspace}
\usepackage{comment}
\usepackage{verbatim}
\usepackage{multirow}
\usepackage{balance}
\usepackage{url}
\usepackage{booktabs}
\usepackage{etoolbox,siunitx}
\usepackage{calc}
\usepackage{pifont,hologo}
\usepackage{color}
\usepackage{ulem}

\newcolumntype{P}[1]{>{\centering\arraybackslash}p{#1}}
\newcolumntype{M}[1]{>{\centering\arraybackslash}m{#1}}

\usepackage[super]{nth}
\usepackage{nicefrac}
\robustify\bfseries

\DeclareMathSymbol{@}{\mathord}{letters}{"3B}

\def\latex/{\LaTeX}
\def\bibtex/{\hologo{BibTeX}}

\makeatletter
\DeclareRobustCommand\onedot{\futurelet\@let@token\@onedot}
\def\@onedot{\ifx\@let@token.\else.\null\fi\xspace}